%% file: main.tex
\pdfoutput=1

\documentclass[11pt]{article}

\usepackage[final]{acl}

\usepackage{times}
\usepackage{latexsym}

\usepackage[T1]{fontenc}

\usepackage[utf8]{inputenc}

\usepackage{microtype}

\usepackage{inconsolata}

\usepackage{graphicx}

\input{math_commands}

\input{macros}

\colorlet{mymax}{gray!0}
\colorlet{mymin}{gray!15}

\usepackage{xcolor}
\usepackage{soul}

\usepackage{float}

\title{State-offset Tuning: State-based Parameter-Efficient Fine-Tuning for State Space Models}

\newcommand*\samethanks[1][\value{footnote}]{\footnotemark[#1]}

\author{
Wonjun Kang$^{1,2}$\thanks{Equal contribution.}
\qquad
Kevin Galim$^{2}$\samethanks
\qquad
Yuchen Zeng$^{3}$\samethanks
\qquad
Minjae Lee$^{2}$
\qquad
\\
\textbf{Hyung Il Koo}$^{2,4}$\thanks{Corresponding author.}
\qquad
\textbf{Nam Ik Cho}$^{1}$
 \\
$^{1}$ Seoul National University \quad $^{2}$ FuriosaAI \quad $^{3}$ UW-Madison \quad $^{4}$ Ajou University \\
{\tt\small \{kangwj1995, kevin.galim, minjae.lee, hikoo\}@furiosa.ai, yzeng58@wisc.edu, nicho@snu.ac.kr}
}

\begin{document}
\maketitle

\input{main/0_abstract}

    \input{figure/figure_teaser_mamba}
\input{main/1_introduction}

\input{main/2_related_works}

\input{main/3_peft_methods}

    \input{table/table_equations}

\input{table/table_views}
    \input{table/table_result_1}

\input{main/4_proposed_method}

\input{main/5_experiments}

\input{main/conclusion}
\input{main/limitations}

\bibliography{custom}

\clearpage
\newpage

\appendix

\section{Visual Comparison of Prompt-based Methods and State-based Methods}
\label{app:detailed_cmp}

\cref{fig:teaser_s6} compares prompt-based methods and state-based methods, including our proposed State-offset Tuning, within the S6 block.

    \input{figure/figure_teaser}

\paragraph{State-based Methods Operate within the SSM Module}

\cref{fig:teaser_s6} shows that prompt-based methods, such as Prefix-Tuning, rely on virtual tokens external to the S6 block. In contrast, state-based methods, such as Initial State Tuning, \stateh{}, and \statey{}, directly adjust state-related features within the S6 block.

\paragraph{{\state} Affects the Current Timestep}

Figure \ref{fig:teaser_s6} illustrates how Prefix-Tuning and Initial State Tuning modify features at early timesteps, indirectly affecting the current state. However, this impact diminishes over time.
In contrast, \stateh{} and \statey{} directly influence the state at each timestep, resulting in more effective adaptation.

\input{main/app_proof}

\section{Low-Rank State-offset Tuning}
\label{app:sec:ours_lr}

\stateh{} shows superior parameter efficiency on Mamba versus other PEFT methods. To further reduce trainable parameters, we can represent the learnable state-offset as a product of two low-rank matrices, inspired by LoRA \citep{hu2021lora}. This is particularly useful for Mamba-2, where the state dimension is larger than in Mamba, leading to an increased number of trainable parameters. In such cases, low-rank techniques can effectively mitigate the parameter overhead. Experimental results of \stateh{} with lower rank on Mamba-2 are provided in \cref{app:mamba2}.

\input{main/app_peft}
\input{main/app_data}

\section{Experimental Details}
\label{app:exp_details}

For every dataset, we select the model size based on how difficult the dataset is and conduct a brief grid search for one epoch using a subset of the data (1k-2k instances) with learning rates of $\set{\num{4e-1}, \num{2e-1}, \num{1e-1}, ..., \num{1e-5}}$. The best learning rate is then selected as the rate that has the lowest training loss. In our experimental results, we report the metric from the best epoch observed on the validation set during training, employing early stopping. Each experiment is conducted once. We apply fine-tuning methods to the SSM module (S6) of Mamba (130M, 1.4B, 2.8B)\footnote{Apache License 2.0, https://huggingface.co/state-spaces/mamba-\{130m,1.4b,2.8b\}} and the SSM module (SSD) of Mamba-2 (130M, 1.3B)\footnote{Apache License 2.0, https://huggingface.co/state-spaces/mamba2-\{130m,1.3b\}} pretrained from Pile~(MIT License, \citet{pile}) using AdamW~\cite{loshchilov2019decoupledweightdecayregularization} with a linear decay schedule for the learning rate. 
In general, we choose hyperparameters for each individual method to ensure that all methods operate within a similar parameter budget. \cref{app:tab:mamba_lr_all,app:tab:peft_hyperparam} show selected learning rates and chosen hyperparameters for each method.
For assessing NLG tasks, we utilize beam search with five beams and a maximum beam length of 1024. BLEU~\citep{bleu}, ROUGE~\citep{rouge}, and METEOR~\citep{meteor} metrics are computed using Hugging Face's evaluate library\footnote{Apache License 2.0, https://huggingface.co/spaces/ evaluate-metric/\{bleu,rouge,meteor\}}.

We use an NVIDIA RTX 3090 24GB for training models with less than 1 billion parameters, and an NVIDIA H100 80GB for larger models. We implemented our project in PyTorch~(Modified BSD license, \citet{pytorch}), utilizing the Hugging Face trainer~(Apache License 2.0, \citet{hftransformers}). We train with batch size 4 for 10 epochs on all datasets except QQP and MNLI for which we use 3 epochs, allowing each training run to finish in under 16 hours. This project spanned three months, utilizing four NVIDIA RTX 3090 24GB GPUs and four NVIDIA H100 80GB GPUs, totaling approximately 17,000 GPU hours.

\begin{table*}[t]
\centering
\resizebox{\linewidth}{!}{
\centering
    \begin{tabular}[t]{lccccccccccccc}
\toprule
\textbf{Model} & \multicolumn{10}{c}{\textbf{Mamba}}  & \multicolumn{3}{c}{\textbf{Mamba-2}} \\
\cmidrule(lr){1-1}\cmidrule(lr){2-11}\cmidrule(lr){12-14}
\textbf{Method / Dataset} & RTE & MRPC & CoLA & SST-2 & QNLI & QQP & MNLI & DART & SAMSum & Spider & DART & SAMSum & Spider \\
\cmidrule(lr){1-1}\cmidrule(lr){2-11}\cmidrule(lr){12-14}
LoRA & 2e-03 & 2e-03 & 4e-05 & 2e-03 & 1e-03 & 1e-03 & 2e-03 & 4e-03 & 2e-03 & 4e-03  & 4e-03 & 2e-03 & 4e-03 \\
Additional-scan & 4e-03 & 2e-03 & 2e-03 & 1e-01 & 2e-03 & 4e-02 & 4e-03 & 4e-03 & 4e-03 & 4e-03 & 2e-02 & 4e-03 & 1e-02 \\
SDT & 1e-03 & 4e-02 & 1e-01 & 4e-02 & 2e-02 & 2e-02 & 1e-01 & 4e-02 & 2e-02 & 4e-02 & - & - & - \\
Initial State Tuning & 4e-04 & 1e-03 & 2e-03 & 2e-03 & 2e-03 & 2e-03 & 2e-03 & 2e-03 & 2e-04 & 1e-03 &4e-03 & 2e-04 & 4e-04 \\
\stateh{} & 1e-03 & 2e-04 & 2e-04 & 1e-04 & 1e-04 & 4e-05 & 4e-04 & 4e-04 & 1e-04 & 2e-04 & 1e-03 & 2e-05 & 2e-05 \\
\stateh{} (low rank) & - & - & - & - & - & - & - & - & - & - & 4e-03 & 2e-04 & 2e-04 \\
\statey{} & 1e-03 & 2e-03 & 1e-03 & 1e-03 & 2e-03 & 1e-03 & 1e-03 & 4e-03 & 1e-03 & 2e-03 & 1e-02 & 2e-04 & 1e-03 \\
\bottomrule
\end{tabular}
}
\caption{Learning rates for each method and dataset. For Mamba and Mamba-2, learning rates for each method and dataset are determined via a small grid search on a dataset subset. The learning rate yielding the best training loss is chosen as the final rate.}
\label{app:tab:mamba_lr_all}
\end{table*}

\begin{table*}[t]
\centering
\resizebox{\linewidth}{!}{
\centering
    \begin{tabular}[t]{lllll}
\toprule
\textbf{Method /Model} & \textbf{Mamba 130M} & \textbf{Mamba 1.4B} & \textbf{Mamba-2 130M} & \textbf{Mamba-2 1.3B} \\
\midrule
LoRA & \makecell[l]{Rank = 8\\$\alpha$ = 8\\Dropout = 0.1\\Modules = all weight matrices in S6} & \makecell[l]{Rank = 8\\$\alpha$ = 8\\Dropout = 0.1\\Modules = all weight matrices in S6} & \makecell[l]{Rank = 16\\$\alpha$ = 16\\Dropout = 0.1\\Modules = all weight matrices in SSD} & \makecell[l]{Rank = 16\\$\alpha$ = 16\\Dropout = 0.1\\Modules = all weight matrices in SSD} \\
\midrule
Additional-scan & \makecell[l]{\#States = 8} & \makecell[l]{\#States = 8} & \makecell[l]{\#States = 32} & \makecell[l]{\#States = 32} \\
\midrule
SDT & \makecell[l]{Freeze \#Channels = 50.0\%\\Freeze \#States = 75.0\%} & \makecell[l]{Freeze \#Channels = 50.0\%\\Freeze \#States = 75.0\%} & - & - \\
\midrule
Initial State Tuning& \makecell[l]{-} & \makecell[l]{-} & \makecell[l]{-} & \makecell[l]{-} \\
\midrule
\stateh{} & \makecell[l]{-} & \makecell[l]{-} & \makecell[l]{-} & \makecell[l]{-} \\
\midrule
\stateh{} (low rank) & - & - & \makecell[l]{Rank = 32} & \makecell[l]{Rank = 64} \\
\midrule
\statey{} & \makecell[l]{-} & \makecell[l]{-} & \makecell[l]{-} & \makecell[l]{-} \\
\bottomrule
\end{tabular}
}
\caption{Hyperparameter settings for each model and PEFT method. In general, we adjust hyperparameters to maintain a similar number of trainable parameters.}
\label{app:tab:peft_hyperparam}
\end{table*}

\section{Additional Experimental Results}
\label{sec:appendix}

\subsection{Mamba Results}

\paragraph{Training Speed and Memory Usage}

We conduct a small experiment to compare the memory usage and training speed of \stateh{} and LoRA, as they performed most similarly in terms of dataset metrics in our experiments. Using a single H100 GPU, we train for 100 batch iterations with a batch size of 4 and a 1K context, continuously measuring memory usage and batch latency.

\cref{app:tab:mem_speed} shows the training speed and maximum memory usage for different Mamba sizes for \stateh{} and LoRA. \stateh{} uses less memory and is faster, even with more trainable parameters. In this experiment, we selected hyperparameters to ensure LoRA has less trainable parameters than \stateh{}. We believe \stateh{}'s efficiency stems from our optimized einsum implementation, enhanced with the opt\_einsum~(MIT License, \citet{opteinsum}) Python package to reduce memory usage and improve latency.

\begin{table}[H]
\centering
\resizebox{\linewidth}{!}{
\centering
\begin{tabular}{rlrrr}
\toprule
\textbf{Model} & \textbf{Method} & \textbf{Params} (\%) & \textbf{Mem.} (GB) & \textbf{Latency} (s) \\
\midrule
\multirow{2}{*}{130M} & \stateh{} & 0.45 & \textbf{4.2} & \textbf{0.13} \\
 & LoRA & 0.35 & 5.44 & 0.18 \\
\midrule
\multirow{2}{*}{370M} & \stateh{} & 0.42 & \textbf{9.36} & \textbf{0.33} \\
 & LoRA & 0.32 & 11.56 & 0.45 \\
\midrule
\multirow{2}{*}{790M} & \stateh{} & 0.3 & \textbf{13.91} & \textbf{0.49} \\
 & LoRA & 0.23 & 17.17 & 0.61 \\
\midrule
\multirow{2}{*}{1.4B} & \stateh{} & 0.23 & \textbf{18.77} & \textbf{0.67} \\
 & LoRA & 0.17 & 22.99 & 0.8 \\
\midrule
\multirow{2}{*}{2.8B} & \stateh{} & 0.19 & \textbf{31.49} & \textbf{1.13} \\
 & LoRA & 0.14 & 37.84 & 1.33 \\
\bottomrule
\end{tabular}
}
\caption{Training speed and memory usage. For each Mamba size, we compare the maximum memory usage and mean latency for processing a single batch during training. Our \stateh{} is compared against LoRA, as it demonstrated the most similar performance in the experiment section. We configure LoRA to use fewer trainable parameters than \stateh{}. Despite this, \stateh{} still consumes less memory and is faster in training.}
\label{app:tab:mem_speed}
\end{table}

\paragraph{FLOP Overhead}

While it is possible to avoid extra FLOP with LoRA in constrained single-task settings by merging weights into the pretrained model, real-world serving scenarios often require a single pretrained model to support multiple downstream tasks simultaneously via multiple LoRA adapters. In such cases, avoiding extra FLOP would require storing separately merged models for each task in memory—an inefficient solution. Alternatively, merging weights dynamically at inference time introduces significant computational bottlenecks. As a result, many recent works focus on serving many LoRA adapters efficiently without weight merging~\citep{s-lora}.

\begin{table}[H]
\centering
\resizebox{\linewidth}{!}{
\centering
    \begin{tabular}{rlrrrrr}
\toprule
\multicolumn{2}{c}{\textbf{Sequence Length}} & \textbf{L=128} & \textbf{L=256}  & \textbf{L=512} & \textbf{L=1024} & \multirow{2.5}{*}{\makecell{\textbf{Relative} \\ (\%)}} \\
\cmidrule(lr){1-2}\cmidrule(lr){3-6}
\multicolumn{1}{c}{\textbf{Model}} & \multicolumn{1}{c}{\textbf{Method}} & \multicolumn{4}{c}{\textbf{GFLOP}} &  \\
\cmidrule(lr){1-1}\cmidrule(lr){2-2}\cmidrule(lr){3-6}\cmidrule(lr){7-7}
\multirow{3}{*}{130M} & Pretrained & 16.45 & 32.90 & 65.81 & 131.61 & 100.000 \\
 & \stateh{} & 16.46 & 32.91 & 65.83 & 131.65 & + 0.029\\
 & LoRA  & 16.61 & 33.21 & 66.42 & 132.84 & + 0.937\\
\midrule
\multirow{3}{*}{370M} & Pretrained & 47.35 & 94.69 & 189.39 & 378.77 & 100.000 \\
 & \stateh{} & 47.36 & 94.72 & 189.44 & 378.87 & + 0.027 \\
 & LoRA & 47.76 & 95.52 & 191.03 & 382.06 & + 0.867 \\
\midrule
\multirow{3}{*}{790M} & Pretrained & 101.22 & 202.44 & 404.88 & 809.75 & 100.000\\
 & \stateh{} & 101.24 & 202.48 & 404.95 & 809.90 & + 0.019 \\
 & LoRA & 101.84 & 203.67 & 407.34 & 814.67 & + 0.608 \\
\midrule
\multirow{3}{*}{1.4B} & Pretrained & 175.23 & 350.45 & 700.90 & 1401.79 & 100.000\\
 & \stateh{} & 175.25 & 350.50 & 701.00 & 1401.99 & + 0.014\\
 & LoRA & 176.05 & 352.09 & 704.17 & 1408.35 & + 0.468 \\
\midrule
\multirow{3}{*}{2.8B} & Pretrained & 353.66 & 707.32 & 1414.63 & 2829.25 & 100.000\\
 & \stateh{} & 353.70 & 707.40 & 1414.80 & 2829.59 & + 0.012\\
 & LoRA  & 355.03 & 710.05 & 1420.09 & 2840.17 & + 0.386 \\
\bottomrule
    \end{tabular}
}
\caption{FLOP overhead across various model sizes and sequence lengths. State-offset Tuning adds less than 0.03\% overhead, whereas LoRA incurs over 30× more extra FLOP compared to ours.}
\label{app:tab:flop}
\end{table}

Given these practical considerations, we evaluate LoRA without weight merging and conduct experiments comparing the additional FLOP of LoRA and our State-offset Tuning method during inference. We use ptflops~\citep{ptflops} to measure computational overhead. 
As shown in \cref{app:tab:flop}, our method adds less than 0.03\% overhead, while LoRA results in more than 30 times the additional FLOP compared to ours. These results highlight the superior FLOP efficiency of our method compared to LoRA.

\paragraph{Mamba 2.8B Results}
\cref{app:spider2} shows the experimental results using Mamba 2.8B. Our \stateh{} outperforms all methods except full fine-tuning.%

\paragraph{Mamba Results on GLUE Dataset}
\cref{tab:app_glue_full} shows the full results on the GLUE dataset using Mamba 130M. Our \stateh{} achieves the highest average score among all PEFT methods.

\subsection{Mamba-2 Results}
\label{app:mamba2}
\cref{tab:app_mamba2} shows experimental results with Mamba-2~\citep{mamba2} models. 
\stateh{} with low-rank adaptation (\cref{app:sec:ours_lr}) significantly reduces the number of trainable parameters. It outperforms existing methods on the Spider benchmark by a large margin and achieves performance comparable to other approaches on the SAMSum and DART datasets.

\input{table/table_results}

\subsection{State-offset Tuning in SSMs vs. Prefix-Tuning in Transformers}

To highlight the effectiveness of State-offset Tuning, we compare its performance with Prefix-Tuning on the Transformer model Pythia~\citep{pythia}. We conduct full fine-tuning and Prefix-Tuning experiments on Pythia 160M on GLUE tasks. The results are shown in \cref{tab:app_pythia}.

Full fine-tuning on Mamba 130M generally surpasses Pythia 160M, consistent with  \citet{gu2023mamba}. Prefix-Tuning on both Mamba and Pythia reaches about 85–90\% of their full fine-tuning performance.

Our State-offset Tuning achieves approximately 98\% of full fine-tuning performance, effectively closing the gap. This success highlights its precise design for SSM-based models.

\begin{table*}[t]
\centering
\resizebox{0.9\linewidth}{!}{
\sisetup{table-auto-round}
\begin{tabular}{cl*{1}{S[table-format=2.2,drop-exponent = true,fixed-exponent = 0,exponent-mode = fixed,]}*{8}{S[table-format=2.1,drop-exponent = true,fixed-exponent = 0,exponent-mode = fixed,]}*{1}{S[table-format=2,drop-exponent = true,fixed-exponent = 0,exponent-mode = fixed,]}}
\toprule
\multicolumn{2}{c}{\textbf{Dataset}} & {\multirow{2.5}{*}{\makecell{\textbf{Params} \\ (\%)}}} & \multicolumn{9}{c}{\textbf{GLUE}}\\
\cmidrule(lr){1-2}\cmidrule(lr){4-12}
\multicolumn{1}{c}{\textbf{Model}} & \multicolumn{1}{c}{\textbf{Method}} & & \textbf{RTE} & \textbf{MRPC} & \textbf{CoLA} & \textbf{SST-2} & \textbf{QNLI} & \textbf{QQP} & \textbf{MNLI} & \textbf{Avg.} & \textbf{Relative (\%)}\\
\cmidrule(lr){1-1}\cmidrule(lr){2-2}\cmidrule(lr){3-3}\cmidrule(lr){4-12}
\multicolumn{1}{c}{\multirow{2}{*}{\makecell{Pythia\\160M}}}& Full Fine-tuning & 100.0 & 64.3 & 77.0 & 20.5 & 88.7 & 85.0 & 88.8 & 79.2 & 71.9 & 100 \\
& Prefix-Tuning & 8.36 & 57.4 & 75.0 & 4.6 & 88.2 & 81.5 & 80.6 & 62.2 & 64.2 & 89 \\
\midrule
\multicolumn{1}{c}{\multirow{3}{*}{\makecell{Mamba\\130M}}}&Full Fine-tuning & 100.00 & 71.1 & 80.6 & 63.2 & 92.2 & 87.4 & 87.9 & 80.8 & 80.5 & 100 \\
& Prefix-Tuning & 22.69 & 67.5 & 75.7 & 43.4 & 91.5 & 83.4 & 83.1 & 35.6 & 68.6 & 85\\
& \textbf{\stateh} & 0.45 & 67.4 & 80.8 & 56.2 & 91.9 & 87.7 & 85.6 & 79.7 & 78.5 & 98 \\
\bottomrule
\end{tabular}}
\caption{Prefix-Tuning experiments on Pythia 160M and Mamba 130M on GLUE tasks. State-offset Tuning for Mamba achieves approximately 98\% of full fine-tuning performance, while Prefix-Tuning reaches about 85–90\% in both SSM and Transformer architectures.
}
\label{tab:app_pythia}
\end{table*}

\subsection{Comparison to Selective Dimension Tuning (SDT)}\label{app:sdt}
We additionally compare our method with Selective Dimension Tuning (SDT)~\citep{galim2024parameter}, a technique derived from theoretical analysis of SSMs.
Note that the hyperparameter selection differs from that used in \citet{galim2024parameter} to ensure the parameter count is more comparable to ours.
As shown in \cref{tab:sdt}, our method consistently outperforms SDT in most cases while using fewer parameters.

\begin{table*}[]
    \centering
    \resizebox{\linewidth}{!}{
    \begin{tabular}{cccccccccccccc}
    \toprule
        \textbf{Model Size} & \multicolumn{9}{c}{\textbf{Mamba 1.4B}} & \multicolumn{4}{c}{\textbf{Mamba 130M}} \\ \cmidrule(lr){1-1} \cmidrule(lr){2-10} \cmidrule(lr){11-14}
        \textbf{Dataset} & \multirow{2.5}{*}{\begin{tabular}{c}
             \textbf{Params}  \\
             (\%)
        \end{tabular}} & \multicolumn{5}{c}{\textbf{Spider}} & \multicolumn{3}{c}{\textbf{SAMSum}} & \multirow{2.5}{*}{\begin{tabular}{c}
             \textbf{Params}  \\
             (\%)
        \end{tabular}}  & \multicolumn{2}{c}{\textbf{DART}} & \textbf{GLUE} \\ \cmidrule(lr){1-1} \cmidrule(lr){3-7} \cmidrule(lr){8-10} \cmidrule(lr){12-13} \cmidrule(lr){14-14}
        \textbf{Method} & & All & Easy & Medium & Hard & Extra & R1 & R2 & RL &  & MET. & BLEU & Avg. \\
        \cmidrule(lr){1-1} \cmidrule(lr){2-2} \cmidrule(lr){3-7} \cmidrule(lr){8-10} \cmidrule(lr){11-11} \cmidrule(lr){12-13} \cmidrule(lr){14-14}
        SDT &  0.26 &19.8& 38.3 & 16.6 & 16.1 & \phantom{0}4.8 & 46.3 & 21.5 & 37.7 & 0.51 & \underline{67.5} & \textbf{48.2} & 63.7 \\
        \textbf{\stateh} & 0.23 & \textbf{57.4} & \textbf{77.4} & \textbf{59.9} & \textbf{44.8} & \textbf{33.7} & \textbf{50.9} & \textbf{26.5} & \textbf{42.4} & 0.45 & \textbf{70.0} & \underline{47.0} & \textbf{78.5} \\
        \textbf{\statey} & 0.01 & \underline{53.0} & \textbf{77.4} & \underline{55.4} & \underline{40.8} & \underline{22.9} & \underline{50.6} & \underline{26.1} & \underline{42.0} & 0.03 & 66.8 & 45.2 & \underline{77.7} \\ \bottomrule
    \end{tabular}}
    \caption{Comparison with Selective Dimension Tuning (SDT)~\citep{galim2024parameter} on Spider, SAMSum, DART, and GLUE.
Our method outperforms SDT in most cases while using fewer parameters.
Note that the hyperparameter configuration of SDT differs from that in \citet{galim2024parameter} to ensure a more comparable parameter count.
    }
    \label{tab:sdt}
\end{table*}

\end{document}

%% file: math_commands.tex
\usepackage{amsmath,amsfonts,bm}

\def\eqref#1{equation~\ref{#1}}

\def\1{\bm{1}}

\def\mI{{\bm{I}}}

\DeclareMathAlphabet{\mathsfit}{\encodingdefault}{\sfdefault}{m}{sl}
\SetMathAlphabet{\mathsfit}{bold}{\encodingdefault}{\sfdefault}{bx}{n}

\def\sR{{\mathbb{R}}}

%% file: macros.tex
\usepackage[utf8]{inputenc} %
\usepackage[T1]{fontenc}    %
\usepackage{url}            %
\usepackage{booktabs}       %
\usepackage{amsfonts}       %
\usepackage{nicefrac}       %
\usepackage{microtype}      %
\usepackage{xcolor}  
\usepackage{multirow}
\usepackage{pmboxdraw}
\usepackage{caption}
\usepackage{subcaption}
\usepackage{tikz, pgfplots}
\usepackage{xcolor}
\usepackage{makecell}
\usepackage{colortbl}

\usepackage{array}
\newcolumntype{H}{>{\setbox0=\hbox\bgroup}c<{\egroup}@{}}

\usepackage{comment}
\usepackage{amsmath}
\usepackage{graphicx}
\usepackage{bm}
\usepackage{nicefrac}
\usepackage{amssymb}
\usepackage{amsthm}
\usepackage{siunitx}

\usepackage{hyperref}
\usepackage[capitalise]{cleveref}
\usepackage{autonum}

\usepackage{wrapfig}
\usepackage[english]{babel}
\usepackage{enumitem} %
\usepackage{pifont}

\def\A{{\bm{A}}}
\def\dA{\overline{\A}}

\def\B{{\bm{B}}}

\def\dB{\overline{\B}}

\def\C{{\bm{C}}}

\def\dt{\bm{\Delta}}

\def\h{\bm{h}}

\def\x{\bm{x}}

\def\Wb{\bm{W}_{\B}}
\def\Wc{\bm{W}_{\C}}
\def\Wdt{\bm{W}_{\dt}}

\def\conv1d{\texttt{Conv1d}}

\def\udh{\widehat{\h}}
\def\udy{\widehat{y}}

\def\dimh{H}

\def\dimd{D}

\def\dimv{V}

\def\s4{\operatorname{S4}}
\def\uds4{\widehat{\s4}}

\def\set#1{\{#1\}}

\colorlet{mygold}{yellow!30}
\colorlet{mysilver}{black!10}

\newcommand{\suf}{_{t+1}}
\newcommand{\pre}{_{\text{prefix}}}

\newcommand{\initial}{Initial State Tuning}
\newcommand{\suffix}{State-offset Tuning}

\newcommand{\state}{State-offset Tuning}
\newcommand{\stateh}{State-offset Tuning ($\h$)}
\newcommand{\statey}{State-offset Tuning ($y$)}

\definecolor{pinegreen}{rgb}{0.0, 0.47, 0.44}
\definecolor{cornellred}{rgb}{0.7, 0.11, 0.11}
\definecolor{cadmiumgreen}{rgb}{0.0, 0.42, 0.24}
\definecolor{royalblue}{rgb}{0.0, 0.14, 0.4}
\definecolor{spirodiscoball}{rgb}{0.06, 0.75, 0.99}
\definecolor{mylightblue}{rgb}{0.85, 0.90, 0.94}
\definecolor{kaistblue}{RGB}{20,135,200}
\definecolor{auburn}{RGB}{166,38,57}

\crefname{appendix}{Sec.}{Secs.}
\Crefname{appendix}{Sec.}{Secs.}
\crefformat{equation}{(#2#1#3)}
\Crefformat{equation}{(#2#1#3)}

\usepackage{tocloft} 
\usepackage{titletoc}

\usepackage{ulem}

\usepackage[linesnumbered,ruled,vlined]{algorithm2e}

\usepackage{environ}
\NewEnviron{revision}{\textcolor{blue}{\BODY}}
\usepackage{comment}

%% file: main/0_abstract.tex
\begin{abstract}
State Space Models (SSMs) have emerged as efficient alternatives to Transformers, mitigating their quadratic computational cost. However, the application of Parameter-Efficient Fine-Tuning (PEFT) methods to SSMs remains largely unexplored. In particular, prompt-based methods like Prompt Tuning and Prefix-Tuning, which are widely used in Transformers, do not perform well on SSMs. To address this, we propose \textit{state-based methods} as a superior alternative to prompt-based methods. This new family of methods naturally stems from the architectural characteristics of SSMs. State-based methods adjust state-related features directly instead of depending on external prompts. Furthermore, we introduce a novel state-based PEFT method: \textit{State-offset Tuning}. At every timestep, our method directly affects the state at the current step, leading to more effective adaptation. Through extensive experiments across diverse datasets, we demonstrate the effectiveness of our method. Code is available at \url{https://github.com/furiosa-ai/ssm-state-tuning}.%
\end{abstract}

%% file: figure/figure_teaser_mamba.tex
\begin{figure}[t]
    \centering
    \includegraphics[width=\linewidth]{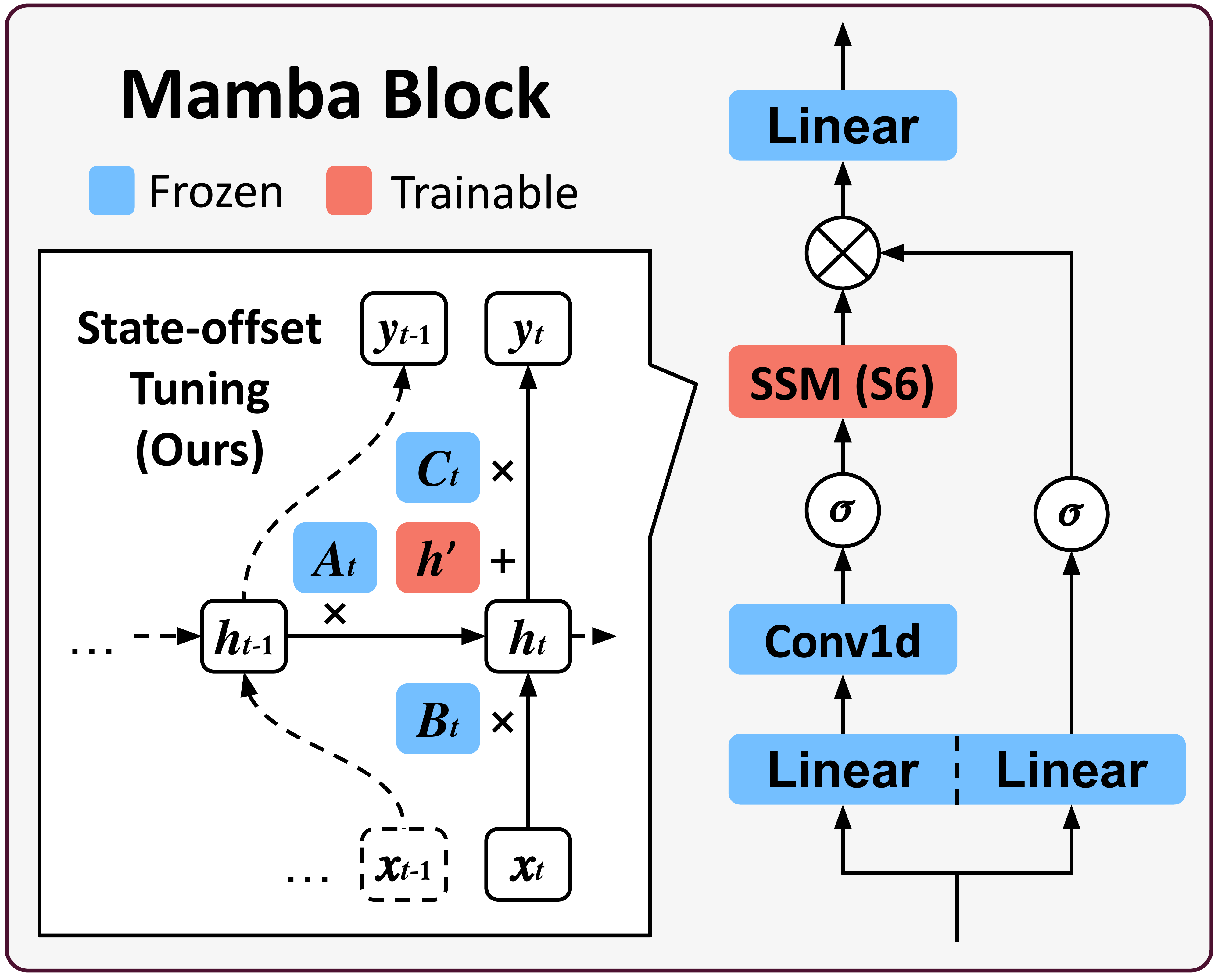}
            \captionsetup{skip=5pt}
    \caption{
         Illustration of our proposed \state{} on a Mamba block~\citep{gu2023mamba}. State-offset Tuning injects a trainable state-offset $\h'$ at each timestep in the SSM module while keeping other parameters frozen, enabling parameter-efficient fine-tuning and improved downstream performance. %
    }
    \label{fig:teaser}
\end{figure}

%% file: main/1_introduction.tex
\section{Introduction}

Large Language Models (LLMs) have gained significant attention for their strong performance in NLP tasks~\citep{achiam2023gpt,brown2020language}, but suffer from the quadratic complexity of Transformer architectures~\citep{vaswani2017attention}.
To mitigate this, subquadratic alternatives have gained interest~\citep{katharopoulos2020transformers,peng2023rwkv,sun2024retnet}, with State Space Models (SSMs) emerging as a promising solution~\citep{gu2023mamba,mamba2}.

Meanwhile, as LLMs scale up, full fine-tuning for downstream tasks becomes prohibitively expensive. Consequently, Parameter-Efficient Fine-Tuning (PEFT)~\citep{houlsby2019parameter,hu2021lora,he2021towards,zaken2021bitfit,liu2021gpt,liu2021p,zeng2023expressive} has emerged, which aims to reduce the number of trainable parameters while achieving adaptation performance comparable to full fine-tuning.

However, research on PEFT methods for SSMs remains limited despite their growing popularity. For instance, prompt-based PEFT methods, such as Prompt Tuning~\citep{lester2021power} and Prefix-Tuning~\citep{li2021prefix}, have been widely applied to Transformers but fail to adapt effectively to SSMs~\citep{galim2024parameter}.
Therefore, new PEFT strategies tailored to SSMs are needed to fully leverage their architectural properties.

To bridge this gap, we introduce \textit{state-based PEFT methods} that leverage the intrinsic properties of SSMs, offering a superior alternative to prompt-based methods.
Building on this concept, we propose \textit{\state{}}. This method directly adjusts the state-related features rather than relying on external prompts, enabling more effective adaptation. %

In summary, our main contributions are:
\begin{itemize}[leftmargin=*, itemsep=3pt, parsep=3pt, topsep=3pt]
\item We introduce \textit{state-based methods}, a new family of PEFT techniques for SSMs, offering a superior alternative to prompt-based approaches.
\item We propose \textit{\state{}} as a new state-based PEFT method.
\item We demonstrate the effectiveness of our method through experiments on a variety of datasets, consistently outperforming existing fine-tuning techniques.
\end{itemize}

%% file: main/2_related_works.tex
\section{Related Works}

\subsection{State Space Models}

Linear State-Space Layers (LSSL) are one of the earliest applications of SSMs in sequence modeling~\citep{gu2021combining}, leveraging HiPPO~\citep{gu2020hippo} to initialize the state matrix. 
However, its high computational overhead limits practicality. \citet{gu2022s4} introduced Structured State Space Models (S4), which mitigate this by structuring the state matrix.
Recently, Mamba \citep{gu2023mamba, mamba2} enhanced modeling capabilities by introducing an input-dependent S6 block.%

\subsection{Parameter-Efficient Fine-Tuning}
In this section, we review existing PEFT methods. For more details, see \cref{app:peft}.

\paragraph{Parameter-based Methods}
One approach to parameter-based PEFT methods is to selectively fine-tune specific layers within the model while keeping the remaining layers frozen. BitFit~\citep{zaken2021bitfit} is a lightweight and effective strategy that focuses solely on fine-tuning a model's bias terms. Furthermore, LoRA~\citep{hu2021lora} represents a notable parameter-based PEFT method by introducing low-rank matrices for weight updates, facilitating efficient adaptation.

\paragraph{Prompt-based Methods}

Instead of fine-tuning model parameters, Prompt Tuning~\citep{lester2021power} enhances models by prepending trainable soft embeddings to the prompt.
Prefix-Tuning~\citep{li2021prefix} builds on this approach by injecting trainable embeddings into each Transformer layer, achieving strong adaptation results for Transformer-based LLMs.

\paragraph{PEFT for SSMs}
Concurrently, ~\citet{galim2024parameter} showed that LoRA outperforms prompt-based methods on SSMs. 
Furthermore, they proposed Selective Dimension Tuning (SDT) for fine-tuning the SSM module while applying LoRA on the linear projection matrices when fine-tuning Mamba models.
~\citet{mambapeft} suggested a new PEFT method called Additional-scan, which increases the hidden state dimension of SSMs, fine-tuning only its additional parameters.

%% file: main/3_peft_methods.tex
\section{PEFT Methods on SSMs}

\paragraph{SSM Preliminaries}
Assuming a single channel dimension, SSMs such as S4~\citep{gu2022s4} transform a signal $x_t \in \sR$ into $y_t \in \sR$ through an $\dimh$-dimensional latent state $\h_t \in \sR^\dimh$ as below:
\begin{align}
\label{eq:rnn_discrete_ssm}
    \h_t = \dA \h_{t-1} + \dB x_{t},&& y_t = \C\h_t,
\end{align}
where $\dB \in \mathbb{R}^{\dimh \times 1}$ controls input influence, $\dA \in \mathbb{R}^{\dimh \times \dimh}$ governs state dynamics, and $\C \in \mathbb{R}^{1 \times \dimh}$ maps the state to the output.
$\dA$ and $\dB$ represent discretized versions of $\A$ and $\B$, parameterized by a learnable step size $\Delta \in \sR$.

In S6 (the SSM module of Mamba), input dependency is integrated by using input-dependent $\dA_t$, $\dB_t$, and $\C_t$ at every timestep.
Specifically, given $D$ channels with $\x_t \in \mathbb{R}^{\dimd}$, learnable parameters $\Wb, \Wc \in \mathbb{R}^{\dimh \times \dimd}$, and $\Wdt \in \mathbb{R}^{\dimd \times \dimd}$ compute $\B_t = \Wb \x_t$, $\C_t = \Wc \x_t$, and $\bm{\Delta} = \Wdt \x_t$.
In this section, we consider S4 for simplicity. %

\subsection{Prompt-based PEFT Methods on SSMs}
\label{subsec:prompt}

\paragraph{Prefix-Tuning Can Update Only the Initial State of an SSM}

Generally, SSMs assume that the initial hidden state is $\h_0=\bm{0}$. We can express $\h_t$ with $\h_0$ as $\h_t\ = \sum\nolimits_{i=1}^t {\dA}^{t-i} \dB_i x_i + {\dA}^t \h_0$.

Assume we have virtual tokens $x_{(-\dimv+1)}, \dots, x_{0}$.
If we prepend virtual tokens as prefix to the input sequence,
we can write the updated $\udh_t$ as below:
\begin{equation}
\udh_t = \h_t + {\dA}^t \sum\nolimits_{i=0}^{\dimv-1} {\dA}^{i} \dB x_{-i} = \h_t + {\dA}^t \h\pre. %
\label{eq:prefix_new}
\end{equation}
By introducing a non-zero $\udh_0$, we can substitute $\udh_0$ for $\h_{\text{prefix}}$ making Prefix-Tuning, or optimizing virtual tokens, equivalent to updating the initial state.
As optimized virtual tokens only affect the initial state $\udh_0$, Prefix-Tuning's expressivity is upper-bounded by updating the initial state directly \cite{galim2024parameter}. 
Since Prefix-Tuning is an extended version of Prompt Tuning, this upper bound is applicable to Prompt Tuning as well.

\citet{galim2024parameter} showed \initial, an advanced version of Prefix-Tuning, which directly optimizes the channel-specific initial state ${\h'}  \in \mathbb{R}^{\dimh}$, resulting in $DH$ trainable parameters in total across all $D$ channels. The updated output $\udy_t$ for Initial State Tuning can be written as in \cref{tab:equations}.

\subsection{State-based Methods: A New Family of PEFT Methods for SSMs}\label{subsec:state}

We define state-based methods as a new family of PEFT methods specifically designed for SSMs. These methods directly modify the intrinsic state-related features within the SSM module.

In contrast, prompt-based methods, such as Prefix-Tuning, influence the hidden state of the SSM module indirectly by introducing external virtual tokens. While both approaches adjust the hidden state of the SSM module, state-based methods operate within the SSM module itself, offering a more direct and expressive adaptation strategy.

Based on our definition, we classify Initial State Tuning as a state-based method. While Initial State Tuning surpasses Prefix-Tuning \citep{galim2024parameter}, it still falls short compared to other finetuning methods on SSMs. To bridge this gap, we propose a novel state-based method for enhanced performance.

%% file: table/table_equations.tex
\begin{table}[H]
\centering
\resizebox{\linewidth}{!}{
\centering
    \begin{tabular}[t]{l|l}
    \toprule
         Initial State Tuning \phantom{\textbf{(Ours)}}& $\udy_t = y_t + \C_t \textstyle{\left( \prod_{i=1}^t \dA_i \right)} \h'$ \\
         \textbf{\stateh{}} & $\udy_t = y_t + \C_t \h'$\\
         \textbf{\statey{}} & $\udy_t = y_t + y'$ \\
        \bottomrule
    \end{tabular}
}
\captionsetup{skip=5pt}
\caption{State-based methods for S6. Our methods eliminate the time-dependent coefficient $\textstyle{\prod_{i=1}^t \dA_i}$, ensuring a uniform effect across timesteps.}
\label{tab:equations}
\end{table}

%% file: table/table_views.tex
\begin{table}[H]
\centering
\resizebox{\linewidth}{!}{
\centering
    \begin{tabular}[t]{l|cc}
    \toprule
         Prompt-based &Timestep $T$ & Timestep $T+1$  \\
    \midrule
         Prefix & \multicolumn{2}{c}{$[\textbf{prefix}, x_{1}, \dots, x_{T}]$ $\rightarrow$ $[\textbf{prefix}, x_{1}, \dots, x_{T}, \x_{T+1}]$} \\
    \midrule
         Suffix & \multicolumn{2}{c}{$[x_{1}, \dots, x_{T}, \textbf{suffix}]$ $\rightarrow$ $[x_{1}, \dots, x_{T},\textbf{suffix},x_{T+1}]$}\\
        \midrule
         Iterative Suffix & \multicolumn{2}{c}{$[x_{1}, \dots, x_{T}, \textbf{suffix}]$ $\rightarrow$ $[x_{1}, \dots, x_{T},x_{T+1},\textbf{suffix}]$}\\
        \bottomrule
    \end{tabular}
}
\captionsetup{skip=5pt}
\caption{Comparison of Prefix-Tuning, Suffix-Tuning, and Iterative Suffix-Tuning.}
\label{tab:views}
\end{table}

%% file: table/table_result_1.tex
\begin{table*}[t]
\centering
\resizebox{\linewidth}{!}{
\sisetup{table-auto-round,mode=text}
\begin{tabular}{cl*{1}{S[table-format=2.2,drop-exponent = true,fixed-exponent = 0,exponent-mode = fixed,]}*{1}{S[table-format=2.1,drop-exponent = true,fixed-exponent = 0,exponent-mode = fixed,]}*{4}{S[table-format=2.1,drop-exponent = true,fixed-exponent = 0,exponent-mode = fixed,]}*{3}{S[table-format=2.1,drop-exponent = true,fixed-exponent = 0,exponent-mode = fixed,]}*{1}{S[table-format=2.2,drop-exponent = true,fixed-exponent = 0,exponent-mode = fixed,]}*{3}{S[table-format=2.1,drop-exponent = true,fixed-exponent = 0,exponent-mode = fixed,]}}
\toprule
\multicolumn{2}{c}{\textbf{Model Size}} &\multicolumn{9}{c}{\textbf{Mamba 1.4B}} & \multicolumn{4}{c}{\textbf{Mamba 130M}}\\
\cmidrule(lr){1-2}\cmidrule(lr){3-11}\cmidrule(lr){12-15}
\multicolumn{2}{c}{\textbf{Dataset}} & {\multirow{2.5}{*}{\makecell{\textbf{Params} \\ (\%)}}} & \multicolumn{5}{c}{\textbf{Spider}} & \multicolumn{3}{c}{\textbf{SAMSum}} &{\multirow{2.5}{*}{\makecell{\textbf{Params} \\ (\%)}}}&\multicolumn{2}{c}{\textbf{DART}}&\multicolumn{1}{c}{\textbf{GLUE}}\\
\cmidrule(lr){1-2}\cmidrule(lr){4-8}\cmidrule(lr){9-11}\cmidrule(lr){13-14}\cmidrule(lr){15-15}
\multicolumn{1}{c}{\textbf{Type}}& \multicolumn{1}{c}{\textbf{Method}} & & {All} & {Easy} & {Medium} & {Hard} & {Extra} & {R1} & {R2} & {RL} & &{MET.} & {BLEU}& {Avg.}\\
\cmidrule(lr){1-1}\cmidrule(lr){2-2}\cmidrule(lr){3-3}\cmidrule(lr){4-11}\cmidrule(lr){12-12}\cmidrule(lr){13-14}\cmidrule(lr){15-15}
\cmidrule(lr){1-1}\cmidrule(lr){2-2}\cmidrule(lr){3-3}\cmidrule(lr){4-11}\cmidrule(lr){12-12}\cmidrule(lr){13-14}\cmidrule(lr){15-15}
\multicolumn{1}{c}{\multirow{3}{*}{-}} & Pretrained & 0.00 & 0.0 & 0.0 & 0.0 & 0.0 & 0.0 & 10.9 & 1.5 & 10.2 & 0.00 & 18.1 & 1.2 & 41.0\\
& Full Fine-tuning (All) &100.0&66.15086793899536&84.27419066429138&69.5067286491394&53.448277711868286&43.37349534034729& 51.197898387908936 & 27.265098690986633 & 42.927393317222595 &100.0&71.00830078125&51.80216431617737&80.5\\
& Full Fine-tuning (S6) &4.456059545468792&\cellcolor{mymax}56.673115491867065&\cellcolor{mymax}76.61290168762207&\cellcolor{mymax}57.847535610198975&45.97701132297516&34.939759969711304&51.13462209701538 &26.891985535621643 & \cellcolor{mymax}42.242348194122314 &4.310565285913944&70.34887075424194&48.67386519908905&79.3\\
\midrule
\multicolumn{1}{c}{\multirow{3}{*}{\makecell{Parameter \\ based}}}&LoRA &0.4643942151277233&\underline{56.3}&\cellcolor{mymax}75.0&\underline{56.5}&\textbf{50.6}&\textbf{33.7}& \cellcolor{mymax}50.52053928375244 & \cellcolor{mymax}26.4 & \underline{42.2} &0.9239127232445363&\underline{69.9}&\textbf{50.8}&\underline{78.3}\\
& BitFit &0.02865633148 & \cellcolor{mymax}51.25725269 & \cellcolor{mymax}74.19354916 & \cellcolor{mymax}50.89685917 & \cellcolor{mymax}43.10344756 & \cellcolor{mymax}26.5060246& \cellcolor{mymax}50.3 & \cellcolor{mymax}25.7 & \cellcolor{mymax}41.9 & \cellcolor{mymax}0.057 & \cellcolor{mymax}67.0 & \cellcolor{mymax}43.7&\cellcolor{mymax}77.9\\
& Additional-scan &0.34269752332618886&26.89&44.35&25.56&21.26&10.24& 37.6& 17.5 & 30.9&0.6804609650495874 & 60.6 & 15.8&\cellcolor{mymax}62.4\\
\cmidrule(lr){1-1}\cmidrule(lr){2-2}\cmidrule(lr){3-3}\cmidrule(lr){4-11}\cmidrule(lr){12-12}\cmidrule(lr){13-15}
\multicolumn{1}{c}{\multirow{2}{*}{\makecell{Prompt \\ based}}} & Prompt Tuning &0.009551198153 & 43.61702204  & 65.32257795  & 42.37668216 & 33.33333433 & 25.30120611& 50.1 & 25.6 & 41.6 &0.03804790468999875&66.18687510490417&39.826393127441406&63.8\\
 & Prefix-Tuning &12.80726891 & 39.65183794 & 65.72580934 & 38.56502175 & 31.03448153 & 15.06024152& \underline{50.6} & \textbf{26.5} & 42.1&22.688043993029535&66.58987998962402&42.462074756622314&68.6\\
\cmidrule(lr){1-1}\cmidrule(lr){2-2}\cmidrule(lr){3-3}\cmidrule(lr){4-11}\cmidrule(lr){12-12}\cmidrule(lr){13-15}
\multicolumn{1}{c}{\multirow{3}{*}{\makecell{State \\ based}}}& Initial State Tuning &0.22872629533389424&51.83752179145813&\textbf{77.8}&51.121073961257935&35.057470202445984&32.5& 50.02278685569763 & 25.960689783096313 & 41.340455412864685 &0.4546719317044869&69.1&46.2&77.4\\
 & \textbf{\stateh{}} &0.22872629533389424&\textbf{57.4}&\underline{77.4}&\textbf{59.9}&\underline{44.8}&\textbf{33.7}&\textbf{50.9}& \textbf{26.5} & \textbf{42.4} &0.4546719317044869&\textbf{70.0}&\underline{47.0}&\textbf{78.5}\\
&\textbf{\statey{}} &0.014326113071832024&53.0&\underline{77.4}&55.4&40.8&22.891566157341003& \underline{50.6} & 26.12304389476776 & 41.97276830673218 &0.028538643106431304&66.82655215263367&45.22758424282074&77.7\\
\bottomrule
\end{tabular}}
\captionsetup{skip=5pt}
\caption{Experimental results for fine-tuning the SSM module (S6) of Mamba~\citep{gu2023mamba} models. We assess Spider and its subsets using execution accuracy, SAMSum with ROUGE-1/2/L scores, DART using METEOR and BLEU scores, and GLUE by calculating the average score. To demonstrate the effectiveness of our methods, we configure the hyperparameters of each method to ensure their parameter budget is comparable to or exceeds that of our methods. \textbf{Bold} and \underline{underline} indicate the best and the second-best results, respectively, among all methods (excluding full fine-tuning). Our \stateh{} outperforms all other methods on most datasets, and our \statey{} shows comparable or better performance than other methods despite its significantly fewer trainable parameters.%
}
\label{tab:results_mamba1}
\end{table*}

%% file: main/4_proposed_method.tex
\section{Proposed State-based PEFT Method}
\label{sec:methods}

In this section, we propose \textit{State-offset Tuning} as a new state-based PEFT method.
A visual comparison with Initial State Tuning and Prefix-Tuning is provided in \cref{app:detailed_cmp}.

 \subsection{\state{}}\label{app:state_based}

{\initial} introduces an additional term $\h'$ with a coefficient ${\dA}^t$ for S4 and $\prod_{i=1}^t \dA_i$ for S6.
However, this coefficient, which varies for each timestep, tends to decrease over time, leading to inconsistent effects. This is related to the issue that SSMs struggle to recall early tokens~\cite{fu2022hungry}. To address this and ensure a consistent effect for each timestep, we introduce \textit{\suffix}, which eliminates this coefficient. 

\suffix\ adds a constant, learnable state-offset $\h'$ to the hidden state $\h$ before obtaining the updated output $\udy_t$ (\cref{fig:teaser}). %
Therefore, unlike {\initial}, {\suffix} does not alter the hidden state dynamics directly. Instead, {\suffix} adds a constant $\h'$ repetitively for each timestep, ensuring a uniform impact.

We formulate \stateh{} for S6 in \cref{tab:equations}, where we optimize ${\h'}  \in \mathbb{R}^{\dimh}$. 
In S4, $\C_t$ does not depend on the input, simplifying to a constant $\C$. This allows us to optimize a bias $y'$ instead of $\h'$ (with ${y'} := \C {\h'}$ for each dimension).   %
We name this method \statey{}. For S4, \statey{} and \stateh{} are equivalent.
In S6, opting for the simpler \statey{} enhances parameter efficiency by decreasing the tunable parameters from $DH$ to $D$.

\subsection{Connection to Prompt-based Methods}
To further validate the methodology of \state{}, we examine its connection to prompt-based methods and demonstrate its correspondence to \textit{Iterative Suffix-Tuning}.

\paragraph{Iterative Suffix-Tuning}

\citet{li2021prefix} showed that in Transformers, inserting virtual tokens at the beginning (Prefix-Tuning) or the end (Suffix-Tuning, referred to as Infix-Tuning in their work) yields similar performance. 

However, for SSMs, the position of the inserted virtual tokens is crucial, as these models tend to forget early tokens.
The effect of Prefix-Tuning and Suffix-Tuning diminishes as the model processes subsequent timesteps. This leads to the question: how can we maintain consistent influence of virtual tokens across all timesteps in SSMs?

To achieve this, we propose Iterative Suffix-Tuning. As shown in \cref{tab:views}, both Prefix-Tuning and Suffix-Tuning hold virtual tokens in fixed positions throughout all timesteps. Conversely, Iterative Suffix-Tuning shifts virtual tokens to the sequence's last position at each timestep, ensuring uniform influence in SSMs. This method is akin to how State-offset Tuning eliminates the time-varying coefficient in Initial State Tuning, enforcing a consistent effect at every timestep. We show that Iterative Suffix-Tuning in SSMs is equivalent to State-offset Tuning (as detailed in \cref{app:suffix_is_state}).

%% file: main/5_experiments.tex
\section{Experiments}

\subsection{Experiment Setup}

We conduct experiments for fine-tuning the SSM module (S6) using pretrained Mamba~\citep{gu2023mamba} and Mamba-2~\citep{mamba2} models on four datasets: Spider~\citep{yu2018spider}, SAMSum~\citep{gliwa2019samsum}, DART~\citep{nan2021dart}, and GLUE~\citep{wang2018glue}.
For further information on datasets, evaluation metrics, and experimental details, refer to \cref{app:datasets,app:exp_details}.
We use LoRA~\citep{hu2021lora}, BitFit~\citep{zaken2021bitfit}, and Additional-scan~\citep{mambapeft}
as parameter-based methods. 
For prompt-based methods, we employ Prompt Tuning~\citep{lester2021power} and Prefix-Tuning\footnote{Following \citet{mambapeft}, our Prefix-Tuning implementation for SSMs corresponds to their Affix-Tuning formulation, adapted specifically for SSM architectures.}~\citep{li2021prefix}. For state-based methods, we utilize Initial State Tuning~\citep{galim2024parameter}, along with our proposed methods, \stateh{} and \statey{}. 

\subsection{Experimental Results}

\cref{tab:results_mamba1} shows the results on Mamba models. 
Additional results, including Mamba-2 results, are provided in \cref{sec:appendix}.
In the appendix, we further compare the training speed, training memory usage, and computational overhead during inference between LoRA and \stateh{}. Our findings show that \stateh{} is faster, more memory-efficient, and introduces lower FLOP overhead compared to LoRA.
Additionally, we evaluate the performance of \stateh{} within SSMs against Prefix-Tuning in Transformers, further highlighting the effectiveness of our approach.

\paragraph{State-based Methods Outperform Prompt-based Methods}

\cref{tab:results_mamba1} shows that all state-based methods outperform prompt-based methods, supporting the claim that state-based methods are superior to prompt-based methods on SSMs.

In particular, our \stateh{} achieves the best results among all tested PEFT methods on most datasets. Our \statey{} outperforms Initial State Tuning on most datasets, using just 0.01\% of the parameters compared to 0.23\% by Initial State Tuning. %

\paragraph{\state{} Outperforms Parameter-Based Methods}

\stateh{} outperforms BitFit across all datasets and surpasses LoRA on most datasets. 
Notably, it also outperforms Additional-scan, a method specifically designed for fine-tuning SSM modules, across all datasets.

Furthermore, \stateh{} achieves performance comparable to full fine-tuning (S6), highlighting the effectiveness of state-based PEFT for SSM modules, despite using significantly fewer parameters. The results from Mamba-2 (\cref{tab:app_mamba2}) further validate the effectiveness of our method.
We also include a comparison to Selective Dimension Tuning (SDT)~\citep{galim2024parameter} in \cref{app:sdt}, showing that our method outperforms SDT while using fewer parameters.

%% file: main/conclusion.tex
\section{Conclusion}

In this paper, we introduce \textit{state-based methods} as a new family of PEFT methods for State Space Models, serving as a superior alternative to prompt-based methods. We propose \textit{\state{}} as a new state-based PEFT method and demonstrate its effectiveness through extensive experiments.

%% file: main/limitations.tex
\section{Limitations}
While we demonstrate that \suffix{} is effective for fine-tuning SSMs in the text domain, its applicability to other domains, such as vision or speech, remains unexplored. Existing PEFT methods, such as LoRA and Prompt Tuning, have been successfully applied across various domains~\citep{jia2022visual,gal2022image,ran2024x}.
Extending \suffix{} to models in other domains, such as Vision Mamba~\citep{zhu2024vision}, is an interesting direction for future work.

\paragraph{Potential Risks}
Our approach enables parameter-efficient fine-tuning (PEFT) of pretrained SSMs, significantly reducing the computational cost of adaptation. While this is beneficial for resource-constrained scenarios, it also presents potential risks. Specifically, adversaries could leverage our method to efficiently fine-tune pretrained SSMs on harmful or biased data, enabling the rapid adaptation of models for malicious purposes with minimal computational resources. This could lead to the proliferation of harmful or deceptive models that reinforce misinformation, bias, or toxicity. To mitigate these risks, future work should explore more robust safety measures, such as integrating ethical fine-tuning constraints and monitoring mechanisms.

%% file: figure/figure_teaser.tex
\begin{figure}[h]
    \centering
    \includegraphics[width=\linewidth]{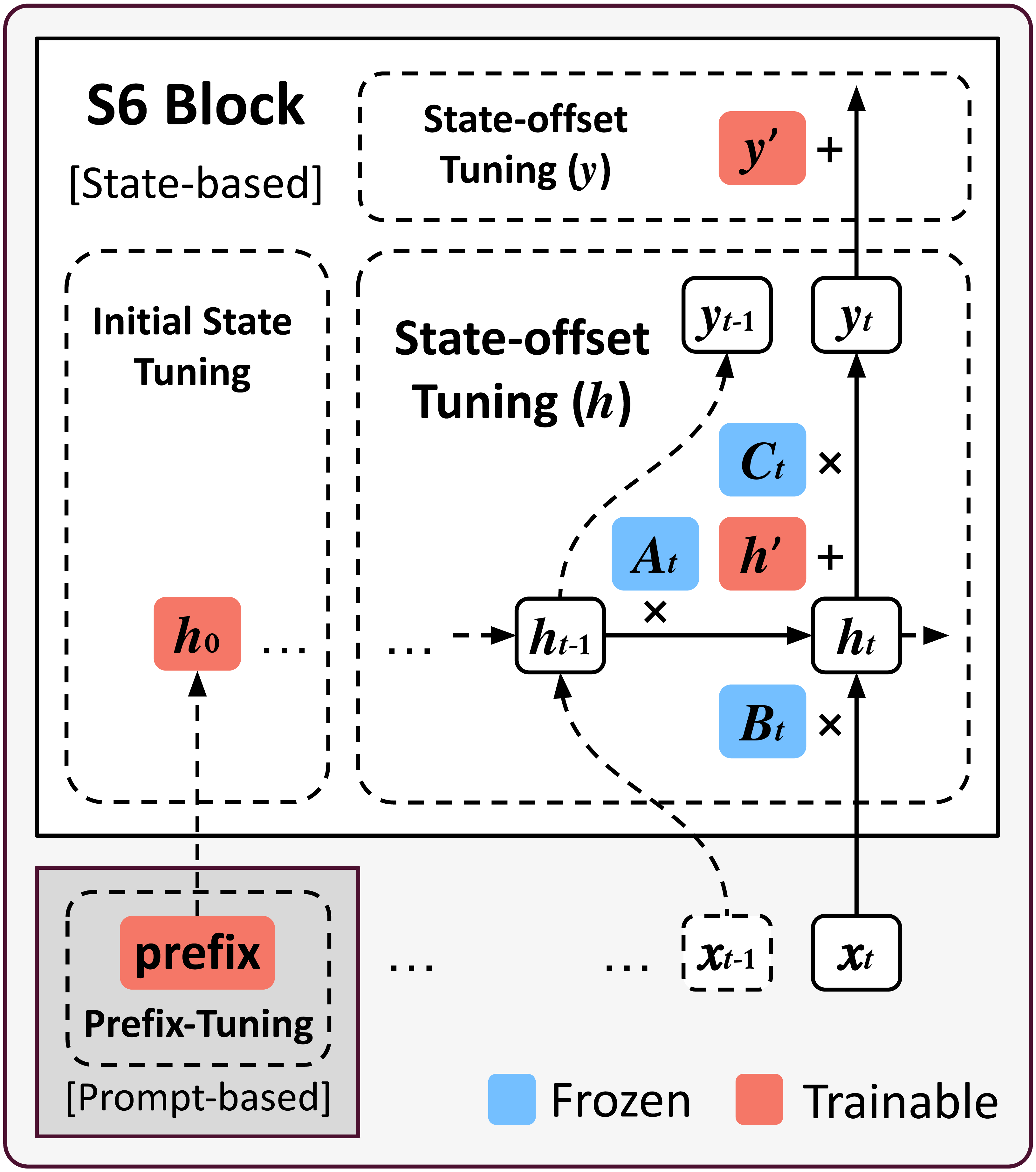}
            \captionsetup{skip=5pt}
    \caption{
         Visual comparison of prompt-based methods and state-based methods in the S6 block. %
    }
    \label{fig:teaser_s6}
\end{figure}

%% file: main/app_proof.tex
\section{Iterative Suffix-Tuning and State-offset Tuning}
\label{app:suffix_is_state}
    \input{figure/figure_views}

In this section, we show that Iterative Suffix-Tuning for SSMs is equivalent to State-offset Tuning.

\paragraph{\suffix{} is Iterative Suffix-Tuning}

\cref{fig:views} provides two different implementations of Iterative Suffix-Tuning on SSMs (S6) with virtual token (suffix) $x_{t+1}$.
\cref{subfig:suffix} views ${t+1}$ as current timestep. In this case, input-dependent $\C_{t+1}=\Wc \x_{t+1}$ is determined solely by the suffix $\x_{t+1} \in \mathbb{R}^{\dimd}$, which is constant at inference time, thus the input dependency of $\C$ is lost, reducing the expressive power of S6.

To address this, we view ${t}$ as current timestep instead and interpret $x_{t+1}$ as future token (\cref{subfig:suffix2}). Consequently, we time-shift $x_{t+1}$ by multiplying it with the inverse of $\dA\suf$.
\begin{align}
\text{\cref{subfig:suffix}:}&\quad y\suf = \C\suf(\dA\suf \h_{t} + \dB\suf x\suf), \\
\text{\cref{subfig:suffix2}:}&\quad y_{t\phantom{+1}} = \bm{C}_t (\h_t + \dA\suf^{-1} \dB\suf x\suf).
\end{align}
Therefore, according to the equation corresponding to \cref{subfig:suffix2}, Iterative Suffix-Tuning can be implemented by updating only $\dA\suf^{-1} \dB\suf x\suf$.
Since this term depends solely on the constant suffix $x_{t+1}$, we can directly replace it with a learnable parameter $\h'$ ($\h' := \dA\suf^{-1} \dB\suf x\suf$), which is equivelant to {\stateh} (\cref{tab:equations}).

%% file: figure/figure_views.tex
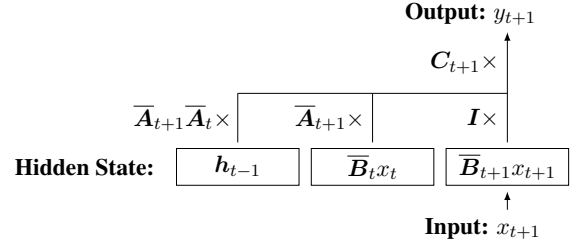
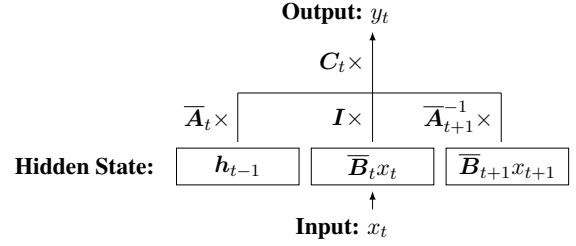
\begin{figure}[ht]
    \centering
    \begin{subfigure}[t]{\linewidth}
        \centering
        \resizebox{\linewidth}{!}{%
    \begin{tikzpicture}
        \path [](-3,0) rectangle  node {\textbf{Hidden State:}} (0,0.6);
        \draw [](0,0) rectangle  node {$\h_{t-1}$} (2,0.6);
        \draw [](2.2,0) rectangle  node {$\dB_t x_t$} (4.2,0.6);
        \draw [](4.4,0) rectangle  node {$\dB\suf x\suf$} (6.4,0.6);
        \node [] at (5,-0.7) {\textbf{Input:} $x\suf$};
        \node [] at (4.8,2.8) {\textbf{Output:} $y\suf$};
        \draw[-latex] (5.4,-0.4) to (5.4,-0.1);
        \draw[-latex] (5.4,0.7) to (5.4,2.5);
        \node [] at (0.1,1.1) {$\dA\suf \dA_t \times$};
        \node [] at (2.5,1.1) {$\dA\suf \times$};
        \node [] at (5,1.1) {$\mI \times$};
        \node [] at (4.7,2) {$\C\suf \times$};
        \draw[] (1,0.7) to (1,1.5);
        \draw[] (3.2,0.7) to (3.2,1.5);
        \draw[] (1,1.5) to (5.4,1.5);
    \end{tikzpicture}
        }
        \caption{Iterative Suffix-Tuning (with $t+1$ as current timestep)}
        \label{subfig:suffix}
        \end{subfigure} 

\medskip

    \begin{subfigure}[t]{\linewidth}
        \centering
        \resizebox{\linewidth}{!}{%
    \begin{tikzpicture}
        \path [](-0.8,0) rectangle  node {\textbf{Hidden State:}} (2.2,0.6);
        \draw [](2.2,0) rectangle  node {$\h_{t-1}$} (4.2,0.6);
        \draw [](4.4,0) rectangle  node {$\dB_t x_t$} (6.4,0.6);
        \draw [](6.6,0) rectangle  node {$\dB\suf x\suf$} (8.6,0.6);
        \node [] at (4.9,-0.7) {\textbf{Input:} $x_t$};
        \node [] at (4.8,2.8) {\textbf{Output:} $y_t$};
        \draw[-latex] (5.4,-0.4) to (5.4,-0.1);
        \draw[-latex] (5.4,0.7) to (5.4,2.5);
        \node [] at (2.7,1.1) {$\dA_t \times$};
        \node [] at (5,1.1) {$\mI \times$};
        \node [] at (4.9,2) {$\C_t \times$};
        \draw[] (3.2,0.7) to (3.2,1.5);
        \draw[] (7.5,0.7) to (7.5,1.5);
        \draw[] (3.2,1.5) to (7.5,1.5);
        \node [] at (6.8,1.1) {$\dA\suf^{-1} \times$}; 
    \end{tikzpicture}
        }
        \caption{Iterative Suffix-Tuning (with $t$ as current timestep)}
        \label{subfig:suffix2}
        \end{subfigure}
        \captionsetup{skip=5pt}
    \caption{Two different implementations of Iterative Suffix-Tuning in S6. We show that \cref{subfig:suffix2} is equivalent to \state{}.}
            \label{fig:views}
    \end{figure}

%% file: main/app_peft.tex
\section{PEFT Baselines}
\label{app:peft}
In this section, we provide a more detailed description of the baseline methods.
\paragraph{LoRA~\citep{hu2021lora}}
LoRA aims to fine-tune large models by maintaining the bulk of pretrained parameters untouched while introducing trainable low-rank matrices within each Transformer's layer. This method leverages linear algebra principles where a large matrix can be effectively approximated by two low-rank matrices, thus reducing the number of parameters. LoRA includes a scaling parameter to adjust the influence of original and LoRA weights during training. 
We use the Hugging Face version~(Apache License 2.0, \citet{hfpeft}) of LoRA for our experiments.

\paragraph{Prompt Tuning~\citep{lester2021power}}
This method involves freezing the entire model and adding a trainable soft prompt to the input. The prompt consists of continuous virtual tokens that provide additional context. %

\paragraph{Prefix-Tuning~\citep{li2021prefix}}
Similar to Prompt Tuning, Prefix-Tuning adds trainable tokens but extends them across every Transformer layer by appending trainable embeddings to the attention matrices. To combat the instability in training these prefixes, an over-parameterized MLP is utilized, which can be discarded after training. %

\paragraph{BitFit~\citep{zaken2021bitfit}}
This PEFT method simplifies fine-tuning by training only the bias terms while freezing the other model weights, drastically reducing trainable parameters. %

\paragraph{SDT~\citep{galim2024parameter}}
SDT (Selective Dimension Tuning) employs a sparse updating approach for the matrices $\A$, $\B$, and $\C$ ($\Wb$ and $\Wc$ for S6), while additionally applying LoRA to the linear projection layers. All remaining layers are kept frozen. The process for determining which parameters to update involves a warmup stage, during which parameters are flagged as updatable if they exhibit a significant gradient magnitude. 
In our SDT experiments, we excluded LoRA from the linear projection layers and focused solely on its S6 component. %

\paragraph{Additional-scan~\citep{mambapeft}}
This approach enhances the model's expressivity by expanding the state dimensions for $\A$, $\Wb$, and $\Wc$. During training, only the added dimensions are marked as trainable. %

%% file: main/app_data.tex
\section{Datasets}\label{app:datasets}

\begin{table}[H]
\centering
\resizebox{\linewidth}{!}{
\centering
    \begin{tabular}[t]{l|rrrrcc}
    \toprule
    Dataset & \#Train & \#Valid & \#Epochs & Model size & Metrics \\
    \midrule
    RTE & 2490 & 277 & 10 & 130m & Acc. \\
    MRPC & 3668 & 408 & 10 & 130m & Acc. \\
    CoLA & 8551 & 1043 & 10 & 130m & Acc. \\
    SST-2 & 67349 & 872 & 10 & 130m & Acc. \\
    QNLI & 104743 & 5463 & 10 & 130m & Acc. \\
    QQP & 363846 & 40430 & 3 & 130m & Acc. \\
    MNLI & 392702 & 19647 & 3 & 130m & Acc. \\
    Spider & 6918 & 1034 & 10 & 1.4B, 2.8B & Acc. \\
    SAMSum & 14732 & 819 & 10 & 1.4B & ROUGE \\
    DART & 62659 & 2768 & 10 & 130m & METEOR, BLEU \\
    \bottomrule
    \end{tabular}
}
\caption{Dataset details. We report the number of training and validation samples, number of training epochs, employed model size and evaluation metrics.}
\label{app:tab:data}
\end{table}

This paper examines four datasets across two domains: Natural Language Understanding (NLU) and Natural Language Generation (NLG). \cref{app:tab:data} presents detailed information for each dataset.

\paragraph{GLUE~\citep{wang2018glue}}
A benchmark comprising nine tasks in English for assessing language understanding models, including sentiment analysis, linguistic acceptability, and question answering. We use the the following datasets: RTE~\cite{gluerte1,gluerte2,gluerte3,gluerte5}, MRPC~\cite{gluemrpc}, CoLA~\citep{gluecola}, SST-2~\citep{gluesst2}, QNLI~\citep{glueqnli}, QQP\footnote{\url{https://quoradata.quora.com/First-Quora-Dataset-Release-Question-Pairs}}, and MNLI~\citep{gluemnli}. Evaluation is mainly through accuracy, except for CoLA where Matthews correlation is used. The final metric is calculated as the average accuracy (Matthews correlation for CoLA) across all datasets. The individual datasets are available under different permissive licenses. We use the version hosted at \url{https://huggingface.co/datasets/nyu-mll/glue}.

\paragraph{SAMSum~\citep{gliwa2019samsum}}
A dataset for dialogue summarization featuring about 16,000 synthetic conversations in English with summaries, created to simulate digital communications with varied tones and styles. Its structure helps in developing systems that process conversational text. The dataset is evaluated via ROUGE score. This dataset is available under the CC BY-NC-ND 4.0 license. We use the version hosted at \url{https://huggingface.co/datasets/Samsung/samsum}.

\paragraph{Spider~\citep{yu2018spider}}
A text-to-SQL dataset with 10,000 annotated SQL queries across 200+ databases, classifying queries from easy to extra hard based on SQL operation complexity. It involves translating English questions to SQL, evaluated via execution accuracy. Execution accuracy considers the output correct if the model's predicted SQL query and the ground truth SQL query yield the same results when executed on the database. This dataset is available under the CC BY-SA 4.0 license. We use the version hosted at \url{https://huggingface.co/datasets/xlangai/spider}.

\paragraph{DART~\citep{nan2021dart}}
Comprising over 80,000 instances, DART focuses on English RDF-to-text generation, organized by structured data triples and corresponding text summaries. It is assessed using METEOR and BLEU metrics. This dataset is available under the MIT license. We use the version hosted at \url{https://huggingface.co/datasets/Yale-LILY/dart}.

%% file: table/table_results.tex
\begin{table*}[t]
\centering
\resizebox{0.7\linewidth}{!}{
\sisetup{table-auto-round}
\begin{tabular}{cl*{1}{S[table-format=2.2,drop-exponent = true,fixed-exponent = 0,exponent-mode = fixed,]}*{1}{S[table-format=2.1,drop-exponent = true,fixed-exponent = 0,exponent-mode = fixed,]}*{4}{S[table-format=2.1,drop-exponent = true,fixed-exponent = 0,exponent-mode = fixed,]}}
\toprule
\multicolumn{2}{c}{\textbf{Model Size}} &\multicolumn{6}{c}{\textbf{Mamba 2.8B}}\\
\cmidrule(lr){1-2}\cmidrule(lr){3-8}
\multicolumn{2}{c}{\textbf{Dataset}} & {\multirow{2.5}{*}{\makecell{\textbf{Params} \\ (\%)}}} & \multicolumn{5}{c}{\textbf{Spider}}\\
\cmidrule(lr){1-2}\cmidrule(lr){4-8}
\multicolumn{1}{c}{\textbf{Type}} & \multicolumn{1}{c}{\textbf{Method}} & & \textbf{All} & \textbf{Easy} & \textbf{Medium} & \textbf{Hard} & \textbf{Extra} \\
\cmidrule(lr){1-1}\cmidrule(lr){2-2}\cmidrule(lr){3-3}\cmidrule(lr){4-8}
\multirow{2}{*}{-}& Full Fine-tuning (All) &100.0&71.76015377044678&87.5&73.54260087013245&63.79310488700867&51.807230710983276\\
& Full Fine-tuning (S6) &4.4387521558002&65.66731333732605&81.85483813285828&68.83407831192017&58.04597735404968&40.963855385780334\\
\midrule
\multicolumn{1}{c}{\multirow{3}{*}{\makecell{Parameter \\ based}}}&LoRA &0.3838049492635724&\underline{63.9}&\underline{86.3}&\textbf{68.2}&49.425286054611206&34.337350726127625\\
& BitFit &0.02367334483 & 59.86460447 & 82.25806355 & 60.76233387 & \textbf{52.9} & 31.32530153\\
& Additional-scan &0.2833 & 35.01 & 61.98 & 31.88 & 27.38 & 12.05\\
\cmidrule(lr){1-1}\cmidrule(lr){2-2}\cmidrule(lr){3-3}\cmidrule(lr){4-8}
\multicolumn{1}{c}{\multirow{2}{*}{\makecell{Prompt \\ based}}}  & Prompt Tuning &0.005917985961427677&50.67698359489441&75.40322542190552&53.811657428741455&37.35632300376892&19.27710771560669\\
& Prefix-Tuning &10.820948510794741&45.06769776344299&75.0&45.06726562976837&32.18390941619873&13.855421543121338\\
\cmidrule(lr){1-1}\cmidrule(lr){2-2}\cmidrule(lr){3-3}\cmidrule(lr){4-8}
\multicolumn{1}{c}{\multirow{3}{*}{\makecell{State \\ based}}}& Initial State Tuning &0.18902876319993944&59.671181440353394&82.2580635547638&62.33183741569519&43.67816150188446&35.54216921329498\\
 & \textbf{\stateh{}} &0.18902876319993944&\textbf{65.0}&\textbf{89.1}&\underline{65.9}&51.724135875701904&\textbf{40.4}\\
 &\textbf{\statey{}} &0.011835271513148709&63.05609345436096&85.88709831237793&64.12556171417236&\underline{52.3}&\underline{37.3}\\
\bottomrule
\end{tabular}}
\caption{Experimental results of fine-tuning the SSM module using pretrained Mamba 2.8B. \stateh{} stands out as the most effective method among all PEFT approaches.
}
\label{app:spider2}
\end{table*}

\begin{table*}[t]
\centering
\resizebox{0.85\linewidth}{!}{
\sisetup{table-auto-round}
\begin{tabular}{cl*{1}{S[table-format=2.2,drop-exponent = true,fixed-exponent = 0,exponent-mode = fixed,]}*{8}{S[table-format=2.1,drop-exponent = true,fixed-exponent = -2,exponent-mode = fixed,]}}
\toprule
\multicolumn{2}{c}{\textbf{Model Size}} &\multicolumn{9}{c}{\textbf{Mamba 130M}}\\
\cmidrule(lr){1-2}\cmidrule(lr){3-11}
\multicolumn{2}{c}{\textbf{Dataset}} & {\multirow{2.5}{*}{\makecell{\textbf{Params} \\ (\%)}}} & \multicolumn{8}{c}{\textbf{GLUE}}\\
\cmidrule(lr){1-2}\cmidrule(lr){4-11}
\multicolumn{1}{c}{\textbf{Type}} & \multicolumn{1}{c}{\textbf{Method}} & & \textbf{RTE} & \textbf{MRPC} & \textbf{CoLA} & \textbf{SST-2} & \textbf{QNLI} & \textbf{QQP} & \textbf{MNLI} & \textbf{Avg.}\\
\cmidrule(lr){1-1}\cmidrule(lr){2-2}\cmidrule(lr){3-3}\cmidrule(lr){4-11}
\multirow{2}{*}{-}& Full Fine-tuning (All) &100.0&0.7111913561820984&0.8063725233078003&0.6319360136985779&0.9220183491706848&0.8740618824958801&0.8787039518356323&0.8076550960540771&0.8045627389635358\\
& Full Fine-tuning (S6) &4.310565285913944&0.6967508792877197&0.7892156839370728&0.5907527804374695&0.9151375889778137&0.8806516528129578&0.8752906322479248&0.8049066066741943&0.7932436891964504\\
\midrule
\multicolumn{1}{c}{\multirow{3}{*}{\makecell{Parameter \\ based}}}&LoRA &0.9239127232445363&0.660649836063385&0.7867646813392639&\textbf{57.8}&0.9082568883895874&\textbf{87.8}&\textbf{86.9}&\textbf{79.8}&\underline{78.3}\\
& BitFit &0.057 & \underline{69.5}& \underline{80.4}& 0.5465259552 & \underline{92.0} & 0.8616145253 & 0.8527826071 & 0.7716699839 & 0.77868355171\\
& Additional-scan &0.68&0.5785&0.7402&0.3862&0.7901&0.7988&0.705&0.369&0.6239714286\\
\cmidrule(lr){1-1}\cmidrule(lr){2-2}\cmidrule(lr){3-3}\cmidrule(lr){4-11}
\multicolumn{1}{c}{\multirow{2}{*}{\makecell{Prompt \\ based}}} & Prompt Tuning &0.04&0.5595667958259583&0.7156862616539001&0.11989180743694305&0.89449542760849&0.7678930759429932&0.7958199381828308&0.6148638129234314&0.6383167313677924\\
 & Prefix-Tuning &22.69&0.675&0.7573529481887817&0.4340396523475647&0.9151375889778137&0.8341570496559143&0.8310660123825073&0.3563902974128723&0.6861762574740818
\\
\cmidrule(lr){1-1}\cmidrule(lr){2-2}\cmidrule(lr){3-3}\cmidrule(lr){4-11}
\multicolumn{1}{c}{\multirow{3}{*}{\makecell{State \\ based}}} & Initial State Tuning &0.4546719317044869&0.667870044708252&0.78432&0.52990&\textbf{92.4}&0.8636280298233032&\underline{86.1}&0.7846490740776062&0.7737\\
 &\textbf{\stateh{}} &0.4546719317044869&0.673913&\textbf{80.8}&\underline{56.2}&0.9185779690742493&\underline{87.7}&0.856&\underline{79.7}&\textbf{78.5}\\
&\textbf{\statey{}} &0.028538643106431304&\textbf{70.0}&0.79560&0.52512&0.9174311757087708&0.8628958463668823&0.856&0.7816969752311707&0.777\\
\bottomrule
\end{tabular}}
\caption{Full results of fine-tuning the SSM module on the GLUE dataset using pretrained Mamba 130M. Our \stateh{} achieves the highest average score among all PEFT methods. %
}
\label{tab:app_glue_full}
\end{table*}

\begin{table*}[t]
\centering
\resizebox{\linewidth}{!}{
\sisetup{table-auto-round,mode=text}
\begin{tabular}{cl*{1}{S[table-format=2.2,drop-exponent = true,fixed-exponent = 0,exponent-mode = fixed,]}*{1}{S[table-format=2.1,drop-exponent = true,fixed-exponent = 0,exponent-mode = fixed,]}*{4}{S[table-format=2.1,drop-exponent = true,fixed-exponent = 0,exponent-mode = fixed,]}*{3}{S[table-format=2.1,drop-exponent = true,fixed-exponent = 0,exponent-mode = fixed,]}*{1}{S[table-format=2.2,drop-exponent = true,fixed-exponent = 0,exponent-mode = fixed,]}*{2}{S[table-format=2.1,drop-exponent = true,fixed-exponent = 0,exponent-mode = fixed,]}}
\toprule
\multicolumn{2}{c}{\textbf{Model Size}} &\multicolumn{9}{c}{\textbf{Mamba-2 1.3B}} & \multicolumn{3}{c}{\textbf{Mamba-2 130M}}\\
\cmidrule(lr){1-2}\cmidrule(lr){3-11}\cmidrule(lr){12-14}
\multicolumn{2}{c}{\textbf{Dataset}} & {\multirow{2.5}{*}{\makecell{\textbf{Params} \\ (\%)}}} & \multicolumn{5}{c}{\textbf{Spider}} & \multicolumn{3}{c}{\textbf{SAMSum}} &{\multirow{2.5}{*}{\makecell{\textbf{Params} \\ (\%)}}}&\multicolumn{2}{c}{\textbf{DART}}\\
\cmidrule(lr){1-2}\cmidrule(lr){4-8}\cmidrule(lr){9-11}\cmidrule(lr){13-14}
\multicolumn{1}{c}{\textbf{Type}}& \multicolumn{1}{c}{\textbf{Method}} & & {All} & {Easy} & {Medium} & {Hard} & {Extra} & {R1} & {R2} & {RL} & &{MET.} & {BLEU}\\
\cmidrule(lr){1-1}\cmidrule(lr){2-2}\cmidrule(lr){3-3}\cmidrule(lr){4-11}\cmidrule(lr){12-12}\cmidrule(lr){13-14}
\multirow{2}{*}{-}& Full Fine-tuning (All) & 100.0 & 64.79690670967102 & 85.88709831237793 & 65.69506525993347 & 54.0229856967926 & 42.16867387294769 &50.95 & 26.89 & 42.45 & 100.0 & 66.5657103061676 & 34.87004339694977 \\
& Full Fine-tuning (SSD) & 2.419408304585285 & 55.12572526931763 & 76.20967626571655 & 56.0538113117218 & 42.52873659133911 & 34.337350726127625 &50.5 & 26.28 & 42.36& 4.169116475966068 & 65.72325229644775 & 39.69655632972717 \\
\midrule
\multicolumn{1}{c}{\multirow{3}{*}{\makecell{Parameter \\ based}}}&LoRA & 0.3681 & 45.36 & 68.95 & 44.39 & 37.36 & 21.08 & 49.67 & 25.91 & 41.66 & 0.7634 & \textbf{70.3} & \textbf{49.6} \\
& BitFit & 0.0158 & 50.87 & 71.37 & 51.57 & \underline{45.4} & 24.1 & \textbf{50.9} & \underline{26.5} & \textbf{42.6} & 0.0338 & 66.18 & 38.98 \\
&Additional-scan  & 0.4671 & 31.91 & 57.26 & 30.49 & 22.99 & 7.23 & 43.0 & 20.13 & 34.82 & 0.9121 & 58.54 & 15.98 \\

\cmidrule(lr){1-1}\cmidrule(lr){2-2}\cmidrule(lr){3-3}\cmidrule(lr){4-11}\cmidrule(lr){12-12}\cmidrule(lr){13-14}
\multicolumn{1}{c}{\multirow{2}{*}{\makecell{Prompt \\ based}}} & Prompt Tuning & 0.0098 & 45.16 & 62.5 & 46.86 & 34.48 & 25.9 & 49.59 & 26.07 & 41.56 & 0.0381 & 65.53 & 36.93 \\
 & Prefix-Tuning & 6.9855 & 47.39 & 70.97 & 48.21 & 32.18 & 25.9 & \underline{50.8} & \underline{26.5} & \textbf{42.6} & 12.8118 & 69.19 & 46.51 \\
\cmidrule(lr){1-1}\cmidrule(lr){2-2}\cmidrule(lr){3-3}\cmidrule(lr){4-11}\cmidrule(lr){12-12}\cmidrule(lr){13-14}

\multicolumn{1}{c}{\multirow{4}{*}{\makecell{State \\ based}}}& Initial State Tuning & 1.8383664749457491 & 54.2553186416626 & 73.38709831237793 & 57.174885272979736 & \underline{45.4} & 27.108433842658997 &50.41 & 26.42 & 42.33& 3.52902152077048 & 65.32499194145203 & 37.22319006919861 \\
 & \textbf{\stateh} & 1.8383664749457491 & \underline{58.5} & \textbf{79.3} & \underline{61.6} & 44.64 & \textbf{33.7} &48.79 & 24.67 & 40.54& 3.52902152077048 & \underline{70.0} & 46.3 \\
 & \textbf{\stateh{} (low rank)}& 0.3499 & \textbf{60.5} & \underline{79.0} & \textbf{65.7} & \textbf{52.3} & \underline{27.7} &50.35 & \textbf{26.8} & 42.5 & 0.7187982591604661 & 69.79 & \underline{47.9} \\
&\textbf{\statey} & 0.014629072996788462 & 43.62 & 66.53 & 42.14 & 36.9 & 21.08 &50.32 & 26.23 & 42.2& 0.02857087586103245 & 65.89 & 38.72 \\
\bottomrule
\end{tabular}}
\caption{Experimental results of fine-tuning the SSM module using pretrained Mamba-2~\citep{mamba2} models. We evaluate Spider and its subsets with execution accuracy, SAMSum using ROUGE-1/2/L scores, and DART through METEOR and BLEU scores. %
\stateh{} with low-rank adaptation (\cref{app:sec:ours_lr}) significantly reduces trainable parameters. It outperforms existing methods on Spider by a wide margin and matches the performance of other approaches on SAMSum and DART.
}
\label{tab:app_mamba2}
\end{table*}